\documentclass{preprint}
\usepackage{booktabs}
\usepackage{graphicx}
\usepackage[colorlinks=true]{hyperref}
\usepackage[numbers,comma,sort&compress]{natbib} 
\usepackage{xcolor}
\usepackage{paralist}
\usepackage{amssymb, amsmath}
\usepackage{colortbl}

\title{A Disease-Aware Dual-Stage Framework for Chest X-ray Report Generation}
\author[1]{Puzhen Wu}
\author[1]{Hexin Dong}
\author[1]{Yi Lin}
\author[2,*]{Yihao Ding}
\author[1,*]{Yifan Peng}
\affil[1]{Population Health Sciences, Weill Cornell Medicine, New York, NY, USA}
\affil[2]{School of Physics, Mathematics and Computing, University of Western Australia, Crawley, Australia}
\affil[*]{Corresponding author(s). Email(s): \url{yip4002@med.cornell.edu}, \url{yihao.ding@uwa.edu.au}}

\begin{document}

\maketitle

\begin{abstract}
Radiology report generation from chest X-rays is an important task in artificial intelligence with the potential to greatly reduce radiologists' workload and shorten patient wait times. Despite recent advances, existing approaches often lack sufficient disease-awareness in visual representations and adequate vision-language alignment to meet the specialized requirements of medical image analysis. As a result, these models usually overlook critical pathological features on chest X-rays and struggle to generate clinically accurate reports. To address these limitations, we propose a novel dual-stage disease-aware framework for chest X-ray report generation. In Stage~1, our model learns Disease-Aware Semantic Tokens (DASTs) corresponding to specific pathology categories through cross-attention mechanisms and multi-label classification, while simultaneously aligning vision and language representations via contrastive learning. In Stage~2, we introduce a Disease-Visual Attention Fusion (DVAF) module to integrate disease-aware representations with visual features, along with a Dual-Modal Similarity Retrieval (DMSR) mechanism that combines visual and disease-specific similarities to retrieve relevant exemplars, providing contextual guidance during report generation. Extensive experiments on benchmark datasets (i.e., CheXpert Plus, IU X-ray, and MIMIC-CXR) demonstrate that our disease-aware framework achieves state-of-the-art performance in chest X-ray report generation, with significant improvements in clinical accuracy and linguistic quality.
\end{abstract}


\section{Introduction}

Generating chest X-ray reports is a critical yet highly complex challenge in medical AI. The goal is to automatically generate comprehensive diagnostic reports from chest radiographs, thereby assisting radiologists by potentially reducing their workload and mitigating workforce shortages. Achieving this goal can also accelerate patient assessment and improve workflow efficiency. 

\begin{figure}[t]
\centering
\includegraphics[width=1.0\linewidth]{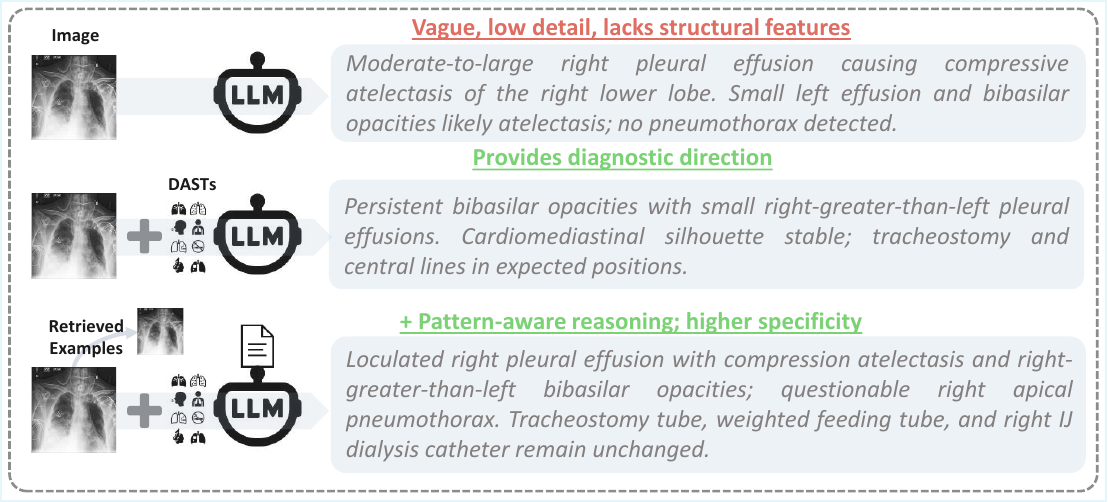}    
\caption{Disease-Aware Semantic Tokens and retrieved examples transform image-only LLM reports from vague to highly specific.}
\label{fig:teaser}
\end{figure}
Most state-of-the-art chest X-ray report generation models follow an encoder–decoder architecture. In this framework, a vision encoder processes the chest X-ray image. At the same time, a language decoder generates the corresponding report. Notably, the R2Gen model introduced a memory-driven Transformer decoder to better capture and retain visual features during the report generation process~\cite{chen2020generating}. Building on this foundation, subsequent works have sought to incorporate domain knowledge and enhance feature representations. For example, the DCL approach by~\citet{li2023dynamic} integrated a dynamic graph of medical entities coupled with contrastive learning to emphasize abnormal findings, thereby demonstrating the benefit of combining structured knowledge with visual encoding. Meanwhile, the adoption of specialized pre-trained language models like Bio-ClinicalBERT~\cite{alsentzer2019publicly} has enriched the inclusion of medical context during text generation. 

The rapid advancement of large language models (LLMs) has further propelled progress in this field. For example, R2GenGPT~\cite{wang2023r2gengpt} aligned extracted visual features with the embedding space of a frozen LLM, effectively leveraging a pre-trained GPT-style model to generate more fluent and clinically accurate reports. Advancements in vision encoding have also played a significant role. Transformer-based encoders, such as ViT~\cite{dosovitskiy2021image}, have been employed to capture global image features, while self-supervised learning methods, such as masked autoencoders (MAE)~\cite{he2022masked}, have improved image representations by pre-training on large collections of unlabeled chest X-rays~\cite{wang2025cxpmrg}. Moreover, multimodal representation learning approaches (e.g., CLIP~\cite{radford2021learning}, which couples images and text with natural language supervision) have inspired techniques to better align visual and textual features for this task. Collectively, these developments have substantially advanced the state of the art in automatic chest X-ray report generation~\cite{xu2025natural}.

Despite substantial progress, several important gaps remain unaddressed. Current models often struggle to capture fine-grained, disease-specific details that are essential in clinical practice. Radiology reports are expected to comprehensively describe all critical findings (e.g., precise locations and severities of lesions or abnormalities). However, standard encoder–decoder models often overlook less prominent findings and tend to generate only high-level impressions. Many existing approaches also lack explicit mechanisms to ensure that each clinically important disease present in the image is both recognized and appropriately described. This can lead to the omission of key conditions or the inclusion of irrelevant observations. Moreover, most contemporary frameworks generate the report solely based on features extracted from the input image and learned language patterns, without utilizing information from similar prior cases that might provide valuable contextual cues. This limitation reduces the model's ability to reinforce its predictions with evidence, particularly for rare conditions or atypical presentations.

In this paper, we propose a novel Disease-Aware Dual-Stage Framework for chest X-ray report generation that directly addresses the issues above. The key innovation lies in incorporating an intermediate semantic understanding stage centered on disease-specific analysis before producing the report. In Stage~1, our model extracts high-level Disease-Aware Semantic Tokens (DASTs) from the image. These DASTs consist of a set of discrete tokens that encode the presence and characteristics of diseases or abnormal findings, thereby creating an explicit semantic bridge between visual features and medical terminology. 

In Stage~2, we leverage a Disease-Visual Attention Fusion (DVAF) module to generate the textual report. This module fuses image features with disease-specific semantic tokens via an attention mechanism. The DVAF module ensures that the decoder attends to the pertinent visual regions and their corresponding disease semantics when constructing each sentence of the report. This approach substantially improves the completeness and accuracy of disease descriptions. Additionally, our framework incorporates a Dual-Modal Similarity Retrieval (DMSR) mechanism that retrieves relevant previous cases by jointly considering image and semantic similarity derived from reports. By referencing image-report pairs similar to the current case, the model gains contextual guidance, helping it avoid common errors and include clinically relevant details that appear in analogous cases. As illustrated in Figure~\ref{fig:teaser}, augmenting an image-only LLM with DASTs and retrieved examples progressively refines the generated report.

We developed and validated our models using three official splits of CheXpert~Plus~\cite{chambon2024chexpert}, IU~X-Ray~\cite{demner2016preparing}, and MIMIC-CXR~\cite{johnson2019mimic}. Extensive experiments demonstrated
that our model achieves state-of-the-art performance across these benchmarks.

To sum up, our main contributions are as follows:
\begin{inparaenum}[(1)]
    \item We introduce Disease-Aware Semantic Tokens (DASTs) that encode disease-specific features, guiding the model toward clinically important abnormalities;
    \item We design a Disease-Visual Attention Fusion (DVAF) module that aligns disease tokens with visual regions, enabling a precise description of findings; 
    \item We develop a Dual-Modal Similarity Retrieval (DMSR) mechanism that retrieves relevant image–report exemplars using visual and semantic similarity, enhancing completeness and factual accuracy; and
    \item We conduct comprehensive evaluations on CheXpert Plus, IU X-Ray, and MIMIC-CXR, achieving state-of-the-art performance across both textual and clinical metrics.
\end{inparaenum}

\section{Related Work}\label{sec:related}

We briefly review three areas of related work: (1) chest X-ray report generation, (2) large-scale pre-trained models for medical vision–language tasks, and (3) state-space models as efficient backbones.

\paragraph{X-ray Medical Report Generation.} 

Early research on automatic report generation for chest X-rays introduced various strategies to improve model performance and coherence. DCL proposed a dynamic graph mechanism on visual features, leveraging external knowledge to strengthen image representations~\cite{li2023dynamic}. RGRG employed an object detector to localize salient lesion regions and generated textual descriptions for each area that were then combined into a complete report~\cite{tanida2023rgrg}. HERGen treated previous reports of the same patient as a temporally ordered sequence, capturing longitudinal information across visits~\cite{xue2024hergen}. More recently, R2GenGPT replaced the traditional decoder with an LLM and substantially improved output quality~\cite{wang2023r2gengpt}. These innovations address domain knowledge, temporal context, and the capabilities of modern LLMs. CXPMRG-Bench~\cite{wang2025cxpmrg} provides the latest controlled experimental results currently through systematic pre-training and benchmark evaluation.

\paragraph{Pre-trained Large Models.} 

With the success of large-scale pre-training, several works have adopted similar ideas for report generation. \citet{xiao2024contextmae} introduced a context-aware masked autoencoder for high-resolution chest X-rays, enabling richer visual features to be extracted from unlabeled images before report generation. CXR-CLIP enlarged the training corpus by synthesizing additional image–text pairs and applied a contrastive loss to align images and reports in a shared space~\cite{you2023cxrclip}. PTUnifier unified different modalities by using learnable visual and textual prompt pools, combining the strengths of fusion- and dual-encoder paradigms~\cite{chen2023ptunifier}. Although promising, these methods commonly rely on Transformer vision encoders that incur quadratic complexity~\cite{dosovitskiy2021image,wu2023improved} and often adopt single-stage pre-training, which limits the use of abundant unpaired X-ray images.

\begin{figure*}[htp]
    \centering
    \includegraphics[width=1.0\linewidth]{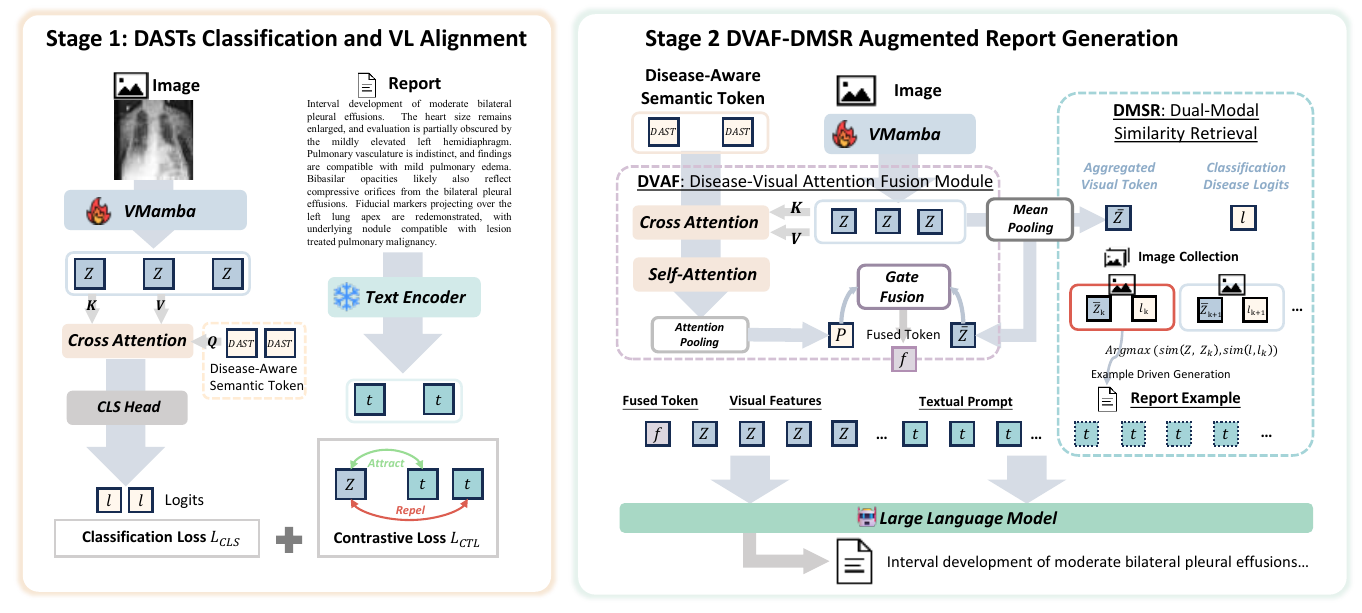}
    \caption{Overview of the proposed two-stage framework. Stage~1 jointly trains a VMamba image encoder and a text encoder to learn disease-aware semantic tokens (DASTs), aligning visual and textual features through classification and contrastive losses. Stage~2 fuses the learned DASTs with visual features, retrieves a similar study via dual-modal similarity retrieval, and feeds these cues into a large language model to generate the final radiology report.}
    \label{fig:net}
\end{figure*}

\paragraph{State-Space Models as Efficient Backbones.}
State-space models have emerged as efficient alternatives to Transformers for sequence modeling because they scale linearly with sequence length. Mamba introduced selective state transitions and a parallelizable recurrence, achieving strong performance with lower computational cost~\cite{gu2024mamba}. Vision Mamba (Vim)~\cite{zhu2024vision} adapted this design to images and provided a global receptive field with linear complexity~\cite{liu2024vmamba}. VMamba extended Vim and has been incorporated into the R2GenCSR generator as a more efficient backbone~\cite{he2023r2gencsr}. Inspired by these successes~\cite{wu2025amd}, we employ a Mamba-based encoder in our dual-stage disease-aware framework, benefiting from faster computation and effective long-range dependency modeling on high-resolution medical images.

\section{Method}

\subsection{Overview}

Our framework integrates three key components (Figure~\ref{fig:net}): a lightweight VMamba encoder for image interpretation~\cite{liu2024vmamba}, a pretrained Bio-ClinicalBERT for medical language modeling~\cite{alsentzer2019publicly}, and an LLM for radiology report generation. This architecture is specifically designed to combine efficient visual representation learning with rich medical language understanding and generative capabilities.

To fully exploit the proposed architecture, we implement a two-stage training strategy. In Stage~1, we introduce 14 DASTs, each corresponding to a specific pathology category. Through cross-attention interactions with visual patch tokens and subsequent multi-label classification tasks, each DAST is trained to encode disease-specific semantic information relevant to its corresponding pathology. We also employ a contrastive loss to more closely align representations learned from visual and textual modalities. 

In Stage~2, we freeze the LLM's parameters and employ a DVAF module to connect the learned disease-aware representations with visual features. Additionally, we introduce a novel DMSR mechanism that jointly considers both visual similarity and disease-specific similarity to retrieve the most relevant exemplar from the training set. This retrieved exemplar provides precise contextual guidance to the report generation process. The final fused visual-disease representations, along with the retrieved exemplar, are then fed into the LLM to generate detailed radiology reports.

\subsection{Stage~1: Disease-Aware Semantic Tokens (DASTs) Classification and Vision Language Alignment}

\paragraph{Visual Encoder.}

To efficiently extract both local anatomical features and global contextual information from chest X-ray images, while maintaining linear computational complexity, we adopt the VMamba architecture~\cite{liu2024vmamba}. VMamba leverages selective state-space models to capture long-range dependencies with $\mathcal{O}(N)$ complexity, offering an advantage over the quadratic $\mathcal{O}(N^2)$ computational cost of conventional Transformers. Given an input image $I\in\mathbb{R}^{H\times W}$, we first divide it into $N$ non-overlapping patches and obtain the initial patch embeddings via
\begin{equation}
    \mathbf{x} = \text{PatchEmbed}(I),
\end{equation}
where $\mathbf{x}\in\mathbb{R}^{N\times C}$ denotes the sequence of raw patch tokens with $C$ feature dimensions.  
These tokens are then fed into $L$ VMamba blocks to model long-range interactions:
\begin{equation}
    \mathbf{z} = \text{VMamba}(\mathbf{x}),
\end{equation}
where $\mathbf{z}\in\mathbb{R}^{N\times C}$ represents the refined visual patch tokens enriched with global contextual information.  

\paragraph{DAST Classification.}

To complement the visual representation, we incorporate $D = 14$ learnable DASTs, denoted as $\{\mathbf{t}_d\}_{d=1}^{D}$, each corresponding to a pathology category defined in the CheXpert ontology~\cite{chambon2024chexpert}. Each DAST $\mathbf{t}_d \in \mathbb{R}^{C}$ is randomly initialized and subsequently optimized during training to encode disease-specific semantics that inform downstream report generation. 

To fuse visual and semantic information, we apply a cross-attention mechanism in which the DASTs serve as queries and visual patch tokens act as keys and values. This design allows each DAST to selectively attend to relevant visual regions associated with its pathology while learning disease-specific semantic representations. The resulting refined disease features are then passed through independent classification heads to enable multi-label pathology prediction within the image.

The learned DASTs capture both visual evidence and medical semantics.
The multi-label classification objective serves as an auxiliary task that encourages DASTs to encode meaningful pathological information while simultaneously guiding the visual encoder to learn discriminative and disease-relevant visual features. This approach ultimately improves the semantic richness and medical accuracy of generated reports in subsequent stages of the framework.

\paragraph{Training Objective.}  

To optimize both disease classification and multimodal alignment, we design a dual loss function that combines a multi-label classification loss with a contrastive alignment loss. This joint training strategy ensures accurate pathology detection and robust vision-language correspondences, which are essential for high-quality report generation in Stage~2.

The classification loss $\mathcal{L}_{\text{CLS}}$ is computed using binary cross-entropy applied to the outputs of the disease-specific classification heads. This auxiliary task encourages the DASTs to encode clinically relevant semantic information, thereby enhancing the medical coherence and accuracy of the generated reports. 

Furthermore, we implement contrastive learning to align visual and textual modalities. This alignment is crucial for bridging the semantic gap between visual pathological findings and their textual descriptions, thereby enabling more effective cross-modal understanding in downstream report generation. Specifically, visual features are extracted from the VMamba encoder via global average pooling over the patch tokens, while textual features are obtained from a frozen Bio-ClinicalBERT encoder. Denote $I_i$ as the $i$-th image and $R_j$ as $j$-th report; the contrastive loss is defined as
\begin{equation}
\mathcal{L}_{\text{CTL}} = \text{Similarity}\left( \text{Visual}(I_i),\text{Textual}(R_j) \right),
\end{equation}
where the similarity is measured by cosine similarity between paired and unpaired samples. The final training objective is formulated as a combination of two loss components
\begin{equation}
\mathcal{L}_{\text{total}} = \mathcal{L}_{\text{CLS}} + \mathcal{L}_{\text{CTL}}.
\end{equation}

This joint optimization ensures that the learned visual representations are both diagnostically informative and semantically aligned with medical language, providing a strong foundation for subsequent report generation.

\subsection{Stage~2: Augmented Report Generation with Disease-Visual Attention Fusion (DVAF) and Dual-Modal Similarity Retrieval (DMSR)}

In Stage~2, we aim to generate detailed radiology reports by leveraging the rich visual features and DASTs learned in Stage~1. We freeze the LLM's parameters and employ DVAF to fuse disease-aware tokens with image features. Simultaneously, DMSR is used to retrieve the most similar case from the training set. The fused features, together with the retrieved exemplar, are then fed into the LLM to generate the report.

\paragraph{DVAF.}  
We begin with the visual patch tokens $\mathbf{z} \in \mathbb{R}^{N \times C}$ and the set of refined DASTs $\{\mathbf{t}_d\}_{d=1}^{14}$, both obtained from Stage~1. To transform this disease-specific information into a representation suitable for report generation, we propose the DVAF module to aggregate the DASTs through a cascade of operations: 
\begin{inparaenum}[(1)]
\item cross-attention with visual patch tokens to inject spatial context, 
\item self-attention to model inter-disease relationships, and 
\item attention pooling to yield a single, unified disease representation
\end{inparaenum}
\begin{equation}
\mathbf{p} = \text{AttnPool}\left(\text{SelfAttn}\left(\text{CrossAttn}\left(\{\mathbf{t}_d\}, \mathbf{z}, \mathbf{z}\right)\right)\right).
\label{eq:dvaf_ops}
\end{equation}
To incorporate global image context, we compute the mean of patch tokens $\bar{\mathbf{z}}$ and fuse it with $\mathbf{p}$ using a gating mechanism
\begin{equation}
\mathbf{f} = \mathbf{W}_{\text{gate}}[\mathbf{p}; \bar{\mathbf{z}}].
\label{eq:gate_fusion}
\end{equation}
The resulting fusion token $\mathbf{f}$ encapsulates both visual and pathological information. This token is then appended to the original patch sequence to construct the final LLM input:
\begin{equation}
\mathbf{V} = [\mathbf{z}_1, \mathbf{z}_2, \dots, \mathbf{z}_N, \mathbf{f}] \in \mathbb{R}^{(N+1) \times C}.
\end{equation}
We then apply a trainable linear projection followed by layer normalization to align this representation with the LLM's input space.

\paragraph{DMSR.}

To improve fluency and completeness of the generated reports, we introduce DMSR. For each input, we construct a composite query vector by combining the mean visual features $\bar{\mathbf{z}}$ and the disease classification logits $\mathbf{l}$. We then compute similarity scores between this query vector and entries stored in the database:
\begin{equation}
s_k = \text{Similarity}(\bar{\mathbf{z}}, \bar{\mathbf{z}}_k) + \lambda \cdot \text{Similarity}(\mathbf{l}, \mathbf{l}_k),
\label{eq:dmsr_similarity}
\end{equation}
where $\lambda$ balances visual and disease similarity. By leveraging both modalities, the DMSR mechanism identifies the most relevant exemplar from the database. The corresponding report $R_{\text{ret}}$ is then retrieved and inserted as an in-context example in the prompt provided to the LLM.

\begin{table*}[t]
\centering
\footnotesize
\setlength{\tabcolsep}{3pt} 
\begin{tabular}{@{}llllcccccc@{}}
\toprule
\textbf{Algorithm} & \textbf{Venue} & \textbf{Encoder} & \textbf{Decoder} & \textbf{BLEU4} & \textbf{ROUGE-L} & \textbf{CIDEr} & \textbf{Precision} & \textbf{Recall} & \textbf{F1} \\
\midrule
R2GenRL~\cite{qin2022reinforced} & ACL'22 & Transformer & Transformer & 0.035 & 0.186 & 0.012 & 0.193 & 0.229 & 0.196 \\
XProNet~\cite{wang2022cross} & ECCV'22 & Transformer & Transformer & 0.100 & 0.265 & 0.121 & 0.314 & 0.247 & 0.259 \\
MSAT~\cite{wang2022msat} & MICCAI'22 & ViT-B/16 & Transformer & 0.036 & 0.156 & 0.018 & 0.044 & 0.142 & 0.057 \\
ORGan~\cite{hou2023organ} & ACL'23 & CNN & Transformer & 0.086 & 0.261 & 0.107 & 0.288 & 0.287 & 0.277 \\
M2KT~\cite{yang2023knowledge} & MIA'21 & CNN & Transformer & 0.078 & 0.247 & 0.077 & 0.044 & 0.142 & 0.058 \\
TIMER~\cite{wu2023timer} & CHIL'23 & Transformer & Transformer & 0.083 & 0.254 & 0.104 & 0.345 & 0.238 & 0.234 \\
R2Gen~\cite{chen2020generating} & EMNLP'20 & Transformer & Transformer & 0.081 & 0.246 & 0.077 & 0.318 & 0.200 & 0.181 \\
R2GenCMN~\cite{chen2021cmn} & ACL'21 & Transformer & Transformer & 0.087 & 0.256 & 0.102 & 0.329 & 0.241 & 0.231 \\
Zhu et al.~\cite{zhu2023longitudinal} & MICCAI'23 & Transformer & Transformer & 0.074 & 0.235 & 0.078 & 0.217 & 0.308 & 0.205 \\
CAMANet~\cite{wang2024camanet} & JBHI'23 & Swin-Former & Transformer & 0.083 & 0.249 & 0.090 & 0.328 & 0.224 & 0.216 \\
ASGMD~\cite{xue2024asgmd} & ESWA'24 & ResNet-101 & Transformer & 0.063 & 0.220 & 0.044 & 0.146 & 0.108 & 0.055 \\
Token-Mixer~\cite{yang2024token} & TMI'23 & ResNet-50 & Transformer & 0.091 & 0.261 & 0.098 & 0.309 & 0.270 & 0.288 \\
PromptMRG~\cite{jin2024promptmrg} & AAAI'24 & ResNet-101 & BERT & 0.095 & 0.222 & 0.044 & 0.258 & 0.265 & 0.281 \\
R2GenGPT~\cite{wang2023r2gengpt} & Meta-Rad'23 & Swin-Former & Llama2 & 0.101 & 0.266 & 0.123 & 0.315 & 0.244 & 0.260 \\
WCL~\cite{yan2021weakly} & EMNLP'21 & Transformer & Transformer & 0.084 & 0.253 & 0.103 & 0.335 & 0.259 & 0.256 \\
VLCI~\cite{chen2023causal} & TIP'25 & Transformer & Transformer & 0.080 & 0.247 & 0.072 & 0.341 & 0.175 & 0.163 \\
MambaXray~\cite{wang2025cxpmrg} & CVPR'25 & Vim & Llama2 & \underline{0.112} & \underline{0.276} & \underline{0.139} &
\underline{0.377} & \underline{0.319} & \underline{0.335} \\
\textit{Ours} & -- & VMamba & Phi-4 & \textbf{0.133} & \textbf{0.291} & \textbf{0.227} & \textbf{0.394} & \textbf{0.356} & \textbf{0.361} \\
\bottomrule
\end{tabular}
\caption{Comparison with state-of-the-art methods on the CheXpert Plus dataset.}
\label{tab:chexpert_results}
\end{table*}

\paragraph{Report Generation and Optimization.}  
The final input to the LLM consists of the retrieved report prompt $R_{\text{ret}}$, the projected visual representation $\tilde{\mathbf{V}}$, and the target report tokens $\mathbf{y}$. The model is trained using a standard language modeling loss
\begin{equation}
\mathcal{L}_{\text{LM}} = -\sum_{t=1}^{T} \log P(y_t \mid y_{<t}, \tilde{\mathbf{V}}, R_{\text{ret}}),
\label{eq:lm_loss}
\end{equation}
where $y_{<t}$ is the previously generated tokens, and $T$ is the report length. Notably, during Stage~2 training, we freeze all components from Stage~1 and only optimize the projection layers and normalization parameters. This approach ensures efficient adaptation to the report generation task while preserving the learned representations from Stage~1.

\section{Experimental Setup}

\subsection{Benchmark Datasets}
We evaluated our approach on three widely used open-access chest X-ray datasets: CheXpert~Plus~\cite{chambon2024chexpert}, IU~X-Ray~\cite{demner2016preparing}, and MIMIC-CXR~\cite{johnson2019mimic}. Each dataset consists of paired chest X-rays and their corresponding reports, as shown in Table \ref{tab:data}. Following established evaluation protocols~\cite{wang2025cxpmrg}, we adopted the same dataset partitioning strategy as in previous works  (Table \ref{tab:data}).

\begin{table}[tbp]
\centering
\small
\begin{tabular}{lrrr}
\toprule
\textbf{Dataset} & \textbf{Training} & \textbf{Val} & \textbf{Testing}\\
\midrule
CheXpert Plus  & 40,463 & 5,780 & 11,562\\
IU X-Ray & 2,069 & 296 & 590\\
MIMIC-CXR & 270,790 & 2,130 & 3,858\\
\bottomrule
\end{tabular}
\caption{Dataset statistics.}
\label{tab:data}
\end{table}

\paragraph{CheXpert Plus.}

CheXpert Plus includes 223,228 chest X-rays, available in both DICOM and PNG formats, paired with 187,711 de-identified radiology reports, each parsed into 11 structured sections. It also encompasses demographic information from 64,725 patients, 14 chest pathology labels, and RadGraph annotations~\cite{jain2021radgraph}. 

\paragraph{IU X-Ray.}

The IU X-ray dataset comprises 7,470 chest X-ray images paired with 3,955 radiology reports, with each report corresponding to either a frontal view or a combination of frontal and lateral view examinations. The reports are structured into four sections: Indication, Comparison, Findings, and Impression. Following common practice in prior work~\cite{chen2020generating}, we concatenated the Findings and Impression sections of each report to form the target text.

\paragraph{MIMIC-CXR.}
MIMIC-CXR consists of chest X-rays and radiology reports from the Beth Israel Deaconess Medical Center Emergency Department in Boston, Massachusetts. Covering the period from 2011 to 2016, it contains 377,110 images and 227,835 reports from 65,379 patients. 

\begin{table*}[t]
\centering
\footnotesize
\setlength{\tabcolsep}{5pt} 
\begin{tabular}{llcccccc}
\toprule
\textbf{Algorithm} & \textbf{Venue} & \textbf{BLEU1} & \textbf{BLEU2} & \textbf{BLEU3} & \textbf{BLEU4} & \textbf{ROUGE-L} & \textbf{CIDEr} \\
\midrule
\rowcolor{gray!30}\textbf{IU X-Ray}&&&&&&&\\
R2Gen~\cite{chen2020generating} & EMNLP'20 & 0.470 & 0.304 & 0.219 & 0.165 & 0.371 & -- \\ 
R2GenCMN~\cite{chen2021cmn} & IJCNLP'21 & 0.475 & 0.309 & 0.222 & 0.170 & 0.375 & -- \\ 
PPKED~\cite{liu2021exploring} & CVPR'21 & 0.483 & 0.315 & 0.224 & 0.168 & 0.376 & 0.351 \\ 
AlignTrans~\cite{you2021align} & MICCAI'21 & 0.484 & 0.313 & 0.225 & 0.173 & 0.379 & -- \\ 
CMCL~\cite{liu2021competence} & ACL'21 & 0.473 & 0.305 & 0.217 & 0.162 & 0.378 & -- \\ 
Clinical-BERT~\cite{yan2022clinicalbert} & AAAI'22 & \underline{0.495} & \textbf{0.330} & 0.231 & 0.170 & 0.376 & 0.432 \\ 
METransformer~\cite{wang2023metransformer} & CVPR'23 & 0.483 & 0.322 & 0.228 & 0.172 & 0.380 & 0.435 \\ 
DCL~\cite{li2023dynamic} & CVPR'23 & -- & -- & -- & 0.163 & 0.383 & \underline{0.586} \\ 
R2GenGPT~\cite{wang2023r2gengpt} & Meta-Rad'23 & 0.465 & 0.299 & 0.214 & 0.161 & 0.376 & 0.542 \\ 
PromptMRG~\cite{jin2024promptmrg} & AAAI'24 & 0.401 & -- & -- & 0.098 & 0.160 & -- \\ 
BootstrappingLLM~\cite{liu2024bootstrapping} & AAAI'24 & \textbf{0.499} & 0.323 & 0.238 & 0.184 & \textbf{0.390} & -- \\ 
MambaXray~\cite{wang2025cxpmrg} & CVPR'25 & 0.491 & \textbf{0.330} & \underline{0.241} & \underline{0.185} & 0.371 & 0.524 \\ 
\textit{Ours} & -- & \underline{0.495} & \underline{0.328} & \textbf{0.242} & \textbf{0.187} & \underline{0.384} & \textbf{0.634} \\ 
\midrule
\rowcolor{gray!30}\textbf{MIMIC-CXR}&&&&&&&\\
R2Gen~\cite{chen2020generating} & EMNLP'20 & 0.353 & 0.218 & 0.145 & 0.103 & 0.277 & -- \\ 
R2GenCMN~\cite{chen2021cmn} & IJCNLP'21 & 0.353 & 0.218 & 0.148 & 0.106 & 0.278 & -- \\ 
PPKED~\cite{liu2021exploring} & CVPR'21 & 0.360 & 0.224 & 0.149 & 0.106 & 0.284 & 0.237 \\ 
AlignTrans~\cite{you2021align} & MICCAI'21 & 0.378 & 0.235 & 0.156 & 0.112 & 0.283 & -- \\ 
CMCL~\cite{liu2021competence} & ACL'21 & 0.344 & 0.217 & 0.140 & 0.097 & 0.281 & -- \\ 
Clinical-BERT~\cite{yan2022clinicalbert} & AAAI'22 & 0.383 & 0.230 & 0.151 & 0.106 & 0.275 & 0.151 \\ 
METransformer~\cite{wang2023metransformer} & CVPR'23 & 0.386 & 0.250 & 0.169 & 0.124 & \textbf{0.291} & \textbf{0.362} \\ 
DCL~\cite{li2023dynamic} & CVPR'23 & -- & -- & -- & 0.109 & 0.284 & \underline{0.281} \\ 
R2GenGPT~\cite{wang2023r2gengpt} & Meta-Rad'23 & 0.408 & 0.256 & 0.174 & 0.125 & 0.285 & 0.244 \\ 
PromptMRG~\cite{jin2024promptmrg} & AAAI'24 & 0.398 & -- & -- & 0.112 & 0.268 & -- \\ 
BootstrappingLLM~\cite{liu2024bootstrapping} & AAAI'24 & 0.402 & 0.262 & 0.180 & 0.128 & \textbf{0.291} & -- \\ 
MambaXray~\cite{wang2025cxpmrg} & CVPR'25 & \underline{0.422} & \underline{0.268} & \underline{0.184} & \textbf{0.133} & \underline{0.289} & 0.241 \\ 
\textit{Ours} & -- & \textbf{0.428} & \textbf{0.272} & \textbf{0.187} & \underline{0.131} & \textbf{0.291} & 0.232 \\ 
\bottomrule
\end{tabular}
\caption{Comparison with state-of-the-art methods on IU X-Ray and MIMIC-CXR.}
\label{tab:iuxray_mimic_results}
\end{table*}

\subsection{Implementation Details}
For Stage~1 training, our training data consists of 480,000 image-text pairs compiled from the training datasets of MIMIC-CXR, CheXpert Plus, and IU X-ray. For all three datasets, the 14-category pathology labels are automatically extracted using the CheXpert labeler~\cite{irvin2019chexpert}, providing a unified supervision signal for Stage~1 classification. The VMamba encoder is initialized with weights pretrained on ImageNet. During this stage, only the visual encoder is set as trainable, and input images are standardized to a resolution of $224 \times 224$ pixels and converted to greyscale. 
For Stage~2 fine-tuning, we evaluated model performance across three datasets with dataset-specific configurations.

Our framework is implemented using PyTorch~\cite{paszke2019pytorch}. 
We utilize the AdamW optimizer~\cite{loshchilov2019adamw} with a learning rate of $1 \times 10^{-4}$. Training incorporates linear warm-up for the first 500 steps, followed by cosine decay scheduling~\cite{loshchilov2017sgdr}. All experiments were conducted on NVIDIA A100 GPUs, using Python 3.10 and PyTorch 2.0.

\subsection{Evaluation Metrics}

\paragraph{Natural Language Generation Metrics.}
We evaluated linguistic quality with BLEU-4~\cite{papineni2002bleu}, ROUGE-L~\cite{lin2004rouge}, and CIDEr~\cite{vedantam2015cider}. These metrics respectively measure n-gram overlap, sequence-level similarity, and term distinctiveness, collectively reflecting lexical accuracy, structural coherence, and semantic fidelity of the generated reports. Here, semantic fidelity is defined as the extent to which the generated report preserves the original clinical meaning.

\paragraph{Clinical Efficacy Metrics.}
To assess diagnostic validity, we applied the CheXpert labeler~\cite{irvin2019chexpert} to extract labels for 14 common chest pathologies from both generated and reference reports. Label agreement was then quantified using Precision, Recall, and F1-score~\cite{powers2011evaluation}, ensuring the generated text accurately captures clinically findings.

\begin{figure}[tbp]
    \centering
    \includegraphics[width=1.0\linewidth]{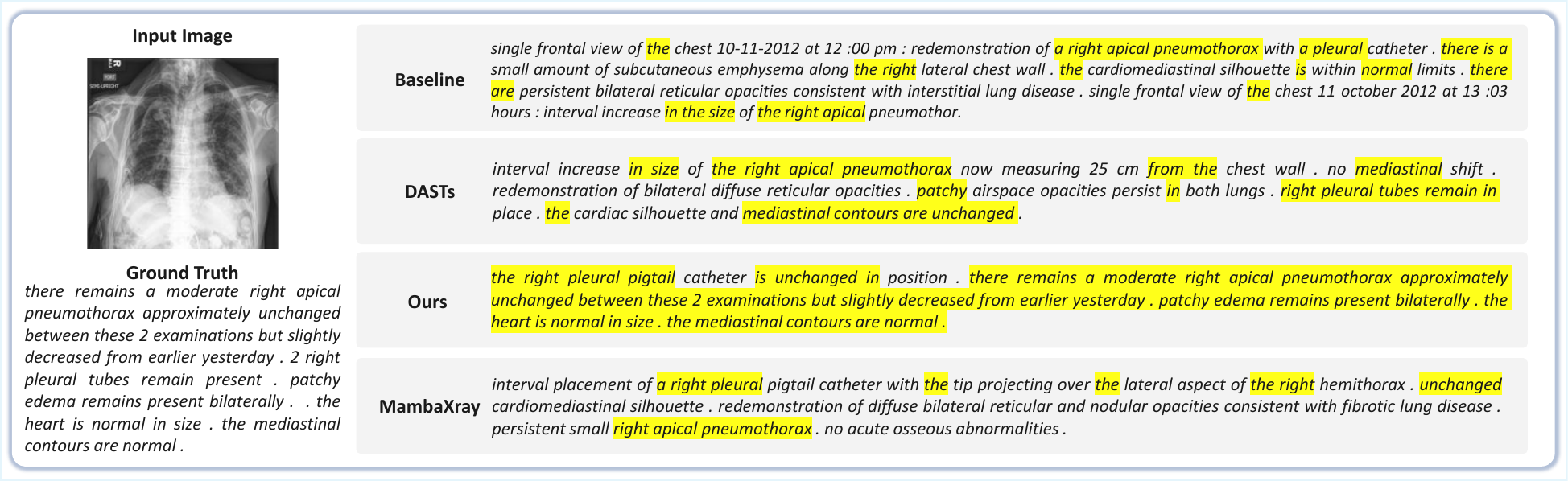}
    \caption{Visualization of generated reports on a sample X-ray.}
    \label{fig:visual}
\end{figure}

\section{Results and Discussion}

\subsection{Results on report generation}

On CheXpert Plus, our framework establishes new state-of-the-art performance benchmarks (Table \ref{tab:chexpert_results}). The method significantly outperforms existing approaches on clinical efficacy metrics~\cite{wang2025cxpmrg}, demonstrating superior diagnostic accuracy and pathology-detection capabilities. The DVAF module effectively integrates visual and disease-specific information, while the DMSR mechanism successfully retrieves relevant exemplars from the large corpus, enabling comprehensive and accurate report generation.

As shown in Table \ref{tab:iuxray_mimic_results}, evaluation on IU X-Ray validates the robustness of our approach across different data scales. Despite limited training data, our method maintains competitive performance. The DAST mechanism effectively captures pathological patterns, and DMSR retrieval enhances report completeness and medical terminology usage, resulting in improved quality scores across linguistic evaluations.

Table \ref{tab:iuxray_mimic_results} also demonstrates that our DVAF-DMSR framework achieves superior performance across NLG metrics compared to existing state-of-the-art methods~\cite{chen2020generating,li2023dynamic,wang2023r2gengpt} in MIMIC-CXR. The integration of DASTs and the DMSR mechanism leads to notable improvements in BLEU-4, ROUGE-L, and CIDEr scores. Our approach consistently outperforms traditional encoder–decoder frameworks and recent vision–language models~\cite{wang2025cxpmrg}, demonstrating the effectiveness of the two-stage training strategy. CIDEr is often lower on MIMIC-CXR because its reports are more free-form and stylistically diverse than those in other datasets with well-structured formats, reducing consistent n-gram overlap despite good BLEU and ROUGE-L scores.

\subsection{Ablation Study}

To validate the effectiveness of each component in our proposed framework, we conduct comprehensive ablation studies on CheXpert Plus and the IU X-Ray datasets. We systematically remove or replace key components to analyze their individual contributions to the overall performance.

\paragraph{Effect of DASTs.}

We first examine the contribution of DASTs by comparing our full model with a baseline that only uses visual patch tokens (Table~\ref{tab:component_ablation}). The results demonstrate that incorporating DASTs significantly improves linguistic quality, validating the importance of explicit disease-aware representations for medical report generation.

\begin{table}[tpb]
\centering
\footnotesize
\setlength{\tabcolsep}{5pt} 
\begin{tabular}{lccc}
\toprule
\textbf{Configuration} & \textbf{BLEU4} & \textbf{ROUGE-L} & \textbf{CIDEr} \\
\midrule
\rowcolor{gray!30}\textbf{CheXpert Plus} & & &\\
Baseline & 0.114 & 0.283 & 0.193 \\
+ DASTs + DVAF & 0.122 & 0.288 & 0.206 \\
+ DASTs + DVAF + DMSR & \textbf{0.133} & \textbf{0.291} & \textbf{0.227} \\
\midrule
\rowcolor{gray!30}\textbf{IU X-Ray} & & &\\
Baseline & 0.161 & 0.374 & 0.550 \\
+ DASTs + DVAF & 0.175 & 0.380 & 0.597 \\
+ DASTs + DVAF + DMSR & \textbf{0.187} & \textbf{0.384} & \textbf{0.634} \\
\bottomrule
\end{tabular}
\caption{Component contribution analysis.}
\label{tab:component_ablation}
\end{table}

\paragraph{Impact of DVAF.}
To assess the effectiveness of our proposed DVAF module, we replace it with simple concatenation or mean pooling of visual and disease tokens (Table~\ref{tab:fusion_ablation}). The comparison reveals that DVAF applies disease-conditioned spatial attention to highlight lesion regions, then fuses this fine-grained focus with global context to produce a unified feature, leading to substantial improvements in report quality.

\begin{table}[tbp]
\centering
\footnotesize
\setlength{\tabcolsep}{5pt} 
\begin{tabular}{lccc}
\toprule
\textbf{Fusion Method} & \textbf{BLEU4} & \textbf{ROUGE-L} & \textbf{CIDEr} \\
\midrule
\rowcolor{gray!30}\textbf{CheXpert Plus} & & &\\
Simple Concatenation & 0.110 & 0.279 & 0.188 \\
Mean Pooling & 0.112 & 0.285 & 0.190 \\
DVAF (Ours) & \textbf{0.122} & \textbf{0.288} & \textbf{0.206}\\
\midrule
\rowcolor{gray!30}\textbf{IU X-Ray} & & &\\
Simple Concatenation & 0.156 & 0.370 & 0.525 \\
Mean Pooling & 0.161 & 0.375 & 0.553 \\
DVAF (Ours) & \textbf{0.175} & \textbf{0.380} & \textbf{0.597} \\
\bottomrule
\end{tabular}
\caption{Comparison of different fusion methods.}
\label{tab:fusion_ablation}
\end{table}

\paragraph{Contribution of DMSR.}

We evaluate the impact of our DMSR mechanism by training models without retrieval augmentation (Table~\ref{tab:component_ablation}). The results show that DMSR consistently enhances performance across all metrics, particularly by improving report completeness and medical terminology accuracy through contextual example guidance. Figure~\ref{fig:visual} qualitatively confirms these findings: the DMSR-enabled model accurately reinstates key phrases, producing a narrative that mirrors the ground truth more faithfully than Baseline and MambaXray\cite{wang2025cxpmrg} outputs.

\subsection{Visualization}

Figure~\ref{fig:visual} provides a visualization of four configurations - Baseline (only visual encoder), DASTs (DASTs+DVAF), our full model, and the SOTA MambaXray - against the ground-truth report for a chest X-ray. Exact phrase matches with the reference are highlighted in yellow. The denser yellow regions in our output demonstrate its higher fidelity and clinical completeness compared to competing methods.

\section{Conclusion}

In this paper, we present a novel framework for automated chest X-ray report generation that introduces Disease-Aware Semantic Tokens (DASTs), a Disease-Visual Attention Fusion (DVAF) module, and a Dual-Modal Similarity Retrieval (DMSR) mechanism. Our two-stage training strategy effectively combines VMamba's computational efficiency with robust visual-semantic alignments and disease-specific representations. Comprehensive experiments across three benchmark datasets demonstrate state-of-the-art performance in both natural language generation and clinical efficacy metrics. Ablation studies confirm the effectiveness of each proposed component, while visualization results validate the clinical relevance of generated reports. Our framework represents a significant advancement toward practical automated radiology reporting systems. The integration of disease-aware semantic guidance, efficient visual encoding, and retrieval-augmented generation provides a promising foundation for future medical AI applications. Future work may explore extending this approach to other medical imaging modalities and clinical deployment scenarios.

\section{Acknowledgments}

This material is based upon work supported by the U.S. National Science Foundation under Award No. CAREER 2145640.

\appendix

\setlength{\bibsep}{3pt plus 0.3ex}
\bibliographystyle{unsrtnat}
\bibliography{ref}

\end{document}